\documentclass[review]{elsarticle}
\usepackage{url}  %Required
\usepackage{graphicx}  %Required
\usepackage[ruled]{algorithm2e} % For algorithms
\usepackage{lineno,hyperref}
\modulolinenumbers[5]
\usepackage{xcolor}
\usepackage{caption}
\usepackage{subfig}
\usepackage{amsthm}
\usepackage{algorithmic}
\usepackage{amsmath}
\usepackage{amssymb}
\usepackage{enumitem}
\usepackage{multirow} 
\newtheorem{theorem}{Theorem}
\journal{Journal of \LaTeX\ Templates}
\journal{Knowledge-based Systems}
\begin{document}
\begin{frontmatter}
\title{Multi-graph Fusion for Multi-view Spectral Clustering}
\author[label1]{Zhao Kang}
\author[label1]{Guoxin Shi}
\author[label1]{Shudong Huang}
\author[label1]{Wenyu Chen}
\author[label1]{Xiaorong Pu}
\author[label2]{Joey Tianyi Zhou}
\author[label1]{Zenglin Xu}

\address[label1]{School of Computer Science and Engineering,\\
University of Electronic Science and Technology of China, \\
Chengdu, Sichuan, 611731, China}
\address[label2]{Institute of High Performance Computing, A*STAR,
Singapore.}
%\[label3]{Corresponding author}
% joey.tianyi.zhou@gmail.com\thanks{Corresponding author}

\begin{abstract}
A panoply of multi-view clustering algorithms has been developed to deal with prevalent multi-view data. Among them, spectral clustering-based methods have drawn much attention and demonstrated promising results recently. Despite progress, there are still two fundamental questions that stay unanswered to date. First, how to fuse different views into one graph. More often than not, the similarities between samples may be manifested differently by different views. Many existing algorithms either simply take the average of multiple views or just learn a common graph. These simple approaches fail to consider the flexible local manifold structures of all views. Hence, the rich heterogeneous information is not fully exploited. Second, how to learn the explicit cluster structure. Most existing methods don't pay attention to the quality of the graphs and perform graph learning and spectral clustering separately. Those unreliable graphs might lead to suboptimal clustering results. To fill these gaps, in this paper, we propose a novel multi-view spectral clustering model which performs graph fusion and spectral clustering simultaneously. The fusion graph approximates the original graph of each individual view but maintains an explicit cluster structure. Experiments on four widely used data sets confirm the superiority of the proposed method.
\end{abstract}

\end{frontmatter}
\section{Introduction}
With the increasing popularity of sensors and multi-camera surveillance systems, one object is often represented from multiple views \cite{chao2019semi,zhu2018multi,tang2018consensus,ding2019multiway}. For example, a person can be uniquely identified in terms of face, fingerprint, iris, and signature; an image can be described by different kinds of descriptors: SIFT, HOG, and LBP, where SIFT is robust to image illumination, noise, and rotation, HOG is sensitive to marginal information, while LBP is a powerful texture feature; the same document can be represented in different languages. Different views can capture distinct perspectives of data. Numerous real-world applications have benefited from multi-view data by leveraging the complementary information \cite{li2017multi,chao2017survey,zhang2018generalized,liu2018late,kang2019multiple}. Thus, multi-view learning has become an important research field \cite{chen2013twkm,huang2019auto}.

As an important ingredient of multi-view learning, multi-view clustering has been widely investigated to identify underlying structures in multi-view data in an unsupervised way \cite{huang2018self,wang2019study}. Although each view contains different fractional information, they together admit the same clustering structure. Simply concatenating all features into a single view and then employing a clustering algorithm on this single view data might not obtain better performance than traditional methods which use single view separately \cite{zhan2018adaptive,huang2019auto}.

In the past decade, plenty of advanced multi-view clustering algorithms have been proposed and they perform effectively by considering the diversity and complementarity of different views. According to the mechanisms on which those methods are based, we can roughly divide them into five categories: co-training style methods \cite{kumar2011cotrain,kumar2011co,tao2018reliable}; multi-kernel learning \cite{liu2017multiple,tzortzis2012kernel,guo2014multiple}; multi-view graph clustering \cite{wang2016iterative,cao2015diversity,gao2015multi,zhan2017graph,wang2017exclusivity,zhang2019multitask}; multi-view subspace clustering \cite{liu2013multi,guo2013convex,xu2017re,liu2018consensus}; multi-task multi-view clustering \cite{zhang2015multi,gu2009learning}.

Among these methods, spectral clustering based multi-view algorithms often report satisfying results. \cite{kumar2011cotrain} proposed a co-training approach to search for the clustering that agree across the views. In this approach, the eigenvectors obtained from one view are used to update the graph of the other view. \cite{kumar2011co} further developed a co-regularized method to look for clustering that are consistent across the views, where the eigenvectors of all views are regularized. Despite their popularity, a common drawback shared by these two methods is that their performance heavily depends on the input graph. It is well-known that small perturbations in the entries of the graph may lead to large perturbations in the eigenvectors, thus leads to inferior clustering accuracy \cite{hunter2010performance,zhao2015automatic,robust2019kang,ding2018semi}. Therefore, constructing an accurate graph is highly desired. 

To this end, a number of graph learning based clustering methods have been proposed recently \cite{kang2017twin,zhang2013graph,kang2019low}. They seek to learn graph from data dynamically. This approach enjoys several nice properties, such as robustness to noise and outliers, independence of similarity metrics. For example, Nie et al. constructed the graph based on adaptive neighbors \cite{nie2014clustering}, i.e., the probability of one data point being the neighbor of another point is treated as a measure of the similarity between them. Afterwards, many researchers extended this idea to deal with multi-view data. \cite{nie2016parameter} reformulated the standard spectral clustering model and put forth a parameter-free multi-view clustering method. This algorithm assumes that all graphs share a common eigenvector. Additionally, this approach takes graph construction and spectral clustering as two separate procedures. As a result, they are not jointly optimized. To solve this problem, \cite{nie2017multi} further developed a unified framework which performs graph learning and spectral clustering simultaneously. However, in this approach, only a common graph is learned based on adaptive neighbors. Consequently, it fails to preserve the flexible local manifold structures for all views which leads to suboptimal clustering performance \cite{wang2016iterative}. In addition, one significant limitation of adaptive neighbors-based graph learning is that it can only capture the intrinsic local structure information of the data.

On the other hand, subspace clustering method has the capability to explore global low-dimensional manifold structure encoded by the data correlations embedded in high-dimensional space \cite{peng2017deep,chen2012fgkm,kang2017kernel,zhang2016joint,li2015robust}. It's based on the self-expressiveness property which assumes that each sample can be linearly represented by the other ones. This representation coefficient matrix $Z$ behaves like the similarity graph matrix \cite{zhang2017latent,xia2014robust,kang2019Clustering,zhang2019robust}. Two widely used assumptions about $Z$ are low-rank \cite{liu2013robust} and sparse \cite{elhamifar2013sparse}. After obtaining the graph, the final clustering result is generated by the spectral clustering algorithm \cite{ng2002spectral,chen2018dnc}. Based on this strategy, varieties of multi-view clustering methods have been proposed. 

Gao et al. \cite{gao2015multi} proposed multi-view subspace clustering algorithm. It learns a graph for each view and enforces a common cluster indicator matrix for all graphs. Thus, the clustering result is consistent for all views. However, this assumption is too strong since the common cluster indicator matrix must negotiate with all graphs. Consequently, the resulted solution might not be optimal. \cite{cao2015diversity} focused on boosting the multi-view clustering by exploring the complementarity of multi-view representations. In specific, they utilize the Hilbert Schmidt Independence Criterion (HSIC) to capture the diversity information. As a result, multiple graphs are built and their average is used as input for spectral clustering. This simple post-processing strategy treats all views equally, which might result in inferior performance. Wang et al. \cite{wang2016iterative} developed a low-rank based multi-view spectral clustering method. Though they added a term to characterize the agreement among the graph, they still used the average of graphs for spectral clustering. This two-step approach might cause unsatisfied results since the averaged graph might not be optimal for subsequent clustering task.
\begin{figure}[!htbp]
\includegraphics[width=.98\textwidth]{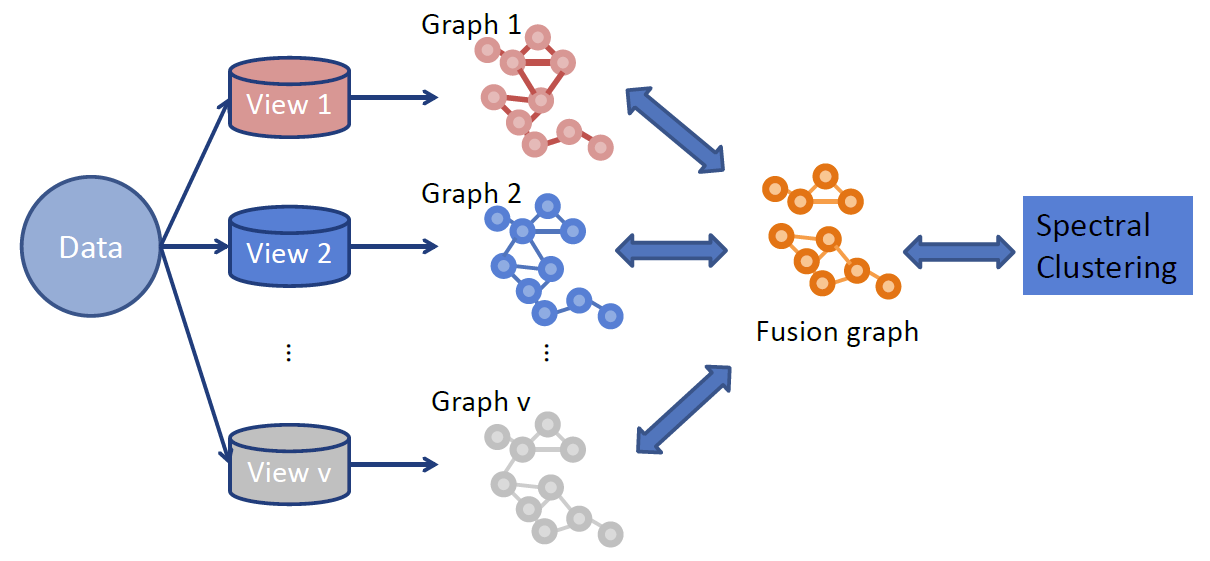}
\caption{Illustration of our GFSC approach. GFSC integrates graph learning, graph fusion, and spectral clustering into a unified framework. The clustering result is further utilized to guide the graph construction and fusion, which in turn contributes to a better clustering. }
\label{flowgraph}
\end{figure}

Despite these progresses, multi-view spectral clustering still arguably faces the following fundamental limitations. First, how to effectively fuse the graphs from all views. Integrating graphs is not trivial since exploration of complementary information of multiple views is the core of multi-view learning \cite{gao2015multi}. Simply taking the average of them fails to consider the discriminative property of views. In many situations, the similarities between samples may be manifested differently by different views. For instance, two video clips that present the same content but in different languages, their audio content will be different. Second, how to consider the explicit cluster structure. It is widely accepted that the clustering results highly depend on the quality of the affinity graph. Many existing methods implement graph construction and spectral clustering separately. Thus, the learned graph might not be ideal for subsequent clustering. 

To solve the above challenging problems, we propose a novel multi-view spectral clustering method which performs graph fusion and spectral clustering simultaneously. Fig. \ref{flowgraph} shows the idea of our approach. The fusion graph approximates the original graph of each individual view but maintains an explicit cluster structure. Experiments on four widely used data sets confirm the superiority of the proposed method. The contributions of this paper are summarized in the following two aspects:
\begin{itemize}
\item A novel graph fusion mechanism is proposed to integrate the multi-view information. It is based on two basic principles: 1) the graph of each view is a perturbation of the consensus graph, and 2) graph that is close to the consensus graph should be assigned a large weight. The graphs are weighted dynamically during the fusion process so that the adversary effect of noise graphs is reduced effectively.
\item The cluster structure of the consensus graph is further considered. As a result, an optimal graph, which has exactly $k$ connected components if there are $k$ clusters, can be readily achieved for clustering. The experimental results confirm its superiority compared to state-of-the-art methods.
\end{itemize} 
\textbf{Notation.} In this paper, matrices are represented by capital letters and vectors are denoted by lower case letters. For an arbitrary matrix $X\in \mathbb{R}^{m\times n}$, its Frobenius norm is $\|X\|_F=\sqrt{\sum_{i=1}^m\sum_{j=1}^n X_{ij}^2}$. The $\ell_2$-norm of vector $x$ is represented by $\|x\|=\sqrt{x^T \cdot x}$, where $T$ means transpose. $Tr(X)$ denotes the trace of $X$. $X\geq 0$ means that all elements of $X$ are nonnegative. $I$ is the identity matrix with a proper size.

\section{Multi-view Spectral Clustering Revisited}
Let $X=\{X^1, X^2, \cdots, X^t\}$ denote the multi-view data with $t$ views. $X^v\in\mathbb{R}^{m_v\times n}(v=1,\cdots,t)$ is the data matrix of view $v$, $m_v$ is the dimension of features in the $v$-th view, and $n$ is the number of samples. Given the adjacent matrix $Z^v\geq 0$ of each view, the graph Laplacian matrix $L^v=D^v-Z^v$, where diagonal matrix $D^v$ is the degree matrix with $d_{ii}^v=\sum_{j=1}^n z_{ij}^v$. 
Assuming that the cluster indicator matrix $F\in\mathbb{R}^{n\times k}$ is the same across all the views, we can formulate the multi-view spectral clustering problem as \cite{gao2015multi}
\begin{equation}
\min_{F,F^TF=I} \sum\limits_{v=1}^t Tr(F^TL^vF), 
\end{equation}
where each graph contributes equally to the final result $F$. In above equation, we ignore the details about the graph construction. Instead of enforcing multiple graphs share the same $F$, several other works simply take the average of graphs and then implement the spectral clustering separately \cite{cao2015diversity,wang2016iterative}. Consequently, the complementary information is not fully exploited since each view is not distinguished from the others. Furthermore, the graphs from different views might differ a lot. It is unrealistic for them to achieve an agreement $F$. Thus, these approaches will lead to inferior clustering result. Some researchers try a linear combination of those graphs \cite{li2015large}. However, the complementary information from multi-view data is not necessarily linearly related. In addition, this linear combination is also sensitive to the weights assigned to each graph. To fill this gap, in this paper, we propose a strategy to integrate the graphs. 

Even if we can achieve a high-quality graph based on our graph fusion principle, we are still unsure whether the graph is suitable for the subsequent clustering task at hand. Ideally, the optimal graph should have an exactly $k$ number of components so that vertices in each connected component of the graph are grouped into the same one cluster. Hence, we go further and incorporate the cluster structure of the consensus graph.

\section{Proposed Multi-graph Fusion for Multi-view Spectral Clustering}
\subsection{Self-expressiveness based Graph Learning}
Self-expressiveness property states that each data sample can be expressed as a linear combination of other samples. This combination coefficient indicates the similarities between samples \cite{vidal2011subspace,liu2013robust}. This similarity graph $Z$ can be obtained by solving 
\begin{equation}
\min_Z \|X-XZ\|_F^2+\alpha \|Z\|_F^2 \quad s.t. \quad Z\geq 0,
\label{single}
\end{equation}
where $\alpha$ is a trade-off parameter. It can be easily extended to multi-view data, i.e.,
\begin{equation}
\min_{Z^v} \sum\limits_{v=1}^t \|X^v-X^vZ^v\|_F^2+\alpha \|Z^v\|_F^2 \quad s.t. \quad Z^v\geq 0,
\label{multi}
\end{equation}
where the same trade-off parameter is often adopted for simplicity. Different graphs capture different aspects of the multi-view data. Then the average of these graphs is often used to achieve the final clustering result \cite{cao2015diversity,wang2016iterative}. That is to say, the consensus graph $S$,
\begin{equation}
S=\frac{\sum\limits_{v=1}^t Z^v}{t},
\end{equation}
is taken as the input for spectral clustering algorithm \cite{ng2002spectral}. It is obvious that this approach fails to distinguish the different contributions of different views. More often than not, some views containing irrelevant or noisy representation might severely damage the graphs and lead to degraded performance. To recap the powerfulness of the complementarity nature of multi-view data, we propose a way to aggregate these basic graphs to form a consensus graph $S$. 
\subsection{Graph Fusion}
Our proposed graph fusion method is based on two intuitive assumptions: 1) the graph $Z^v$ of each view is a perturbation of the consensus graph $S$, and 2) the graph that is close to the consensus graph should be assigned a large weight. The consensus graph $S$ is supposed to capture the ground-truth sample similarity hidden in the multi-view data. To avoid the influence of low quality (noisy) views, we try to assign different weights to different graphs. As a result, we can reach a better clustering performance based on $S$ than that of $Z^v$. 

Based on above principles, our graph fusion mechanism can be formulated as
\begin{equation}
\sum\limits_{v=1}^t w_v\|Z^v-S\|_F^2,
\label{fuse}
\end{equation}
where the weight $w_v$ characterizes the importance of view $v$. We can simply adopt the inverse distance weighting scheme \cite{nie2016parameter,nie2017self}, i.e.,
\begin{equation}
w_v=\frac{1}{2\|Z^v-S\|_F}.
\label{weightw}
\end{equation}
Since $S$ is unknown beforehand, we can calculate it approximately based on an iterative approach. Then we combine Eq. (\ref{fuse}) and (\ref{multi}). It yields 
\begin{equation}
\begin{split}
\min_{Z^v,S}& \sum\limits_{v=1}^t \|X^v-X^vZ^v\|_F^2+\alpha \|Z^v\|_F^2+\beta w_v\|Z^v-S\|_F^2 \\
& s.t. \quad Z^v\geq 0,
\label{firstobj}
\end{split}
\end{equation}
Through solving this problem, we can obtain both the graph for each view and the consensus graph adaptively. Additionally, the graphs are weighted dynamically during the fusion process so that the adversary effect of noise graphs is reduced effectively. Although we can directly implement spectral clustering based $S$, we move forward and consider the cluster structure of it since the current graph $S$ might not be optimal for the subsequent clustering task.

\subsection{Structured Graph Learning }
Ideally, the solution $S$ of problem (\ref{firstobj}) should have exact $k$ connected components, i.e., the data points are already clustered into $k$ clusters. However, the current solution can hardly satisfy to such a condition. This can be fulfilled based on the following theorem \cite{mohar1991laplacian}:
\begin{theorem}
The number of connected components $k$ of the graph $S$ is equal to the multiplicity of zero eigenvalues of its Laplacian matrix $L$. 
\end{theorem}
Since $L$ is a positive semi-definite matrix, its eigenvalues $\sigma_n\geq \cdots\geq\sigma_2\geq \sigma_1\geq0$. Theorem 1 means that if $\sum_{i=1}^k \sigma_i=0$, then our expectation can be approximately satisfied. Hence, we can minimize $\sum_{i=1}^k \sigma_i$ instead to satisfy the requirement. According to Ky Fan's theorem \cite{fan1949theorem}, we can obtain an objective function 
\begin{equation}
\sum\limits_{i=1}^k \sigma_i=\min_{F,F^TF=I} Tr(F^TLF).
\label{fan}
\end{equation}
The right part of this equation is nothing but the objective function of spectral clustering. Hence, Eq. (\ref{fan}) establishes the connection between our requirement for the graph structure and spectral clustering. 

Minimizing Eq. (\ref{fan}), we can approximately guarantee the structure of graph $S$. Therefore, we can combine Eqs. (\ref{fan}) and (\ref{firstobj}) to a single objective function, which fulfills the tasks of graph learning, graph fusion, and spectral clustering. Consequently, our proposed multi-Graph Fusion for multi-view Spectral Clustering (GFSC) can be formulated as
\begin{equation}
\begin{split}
\min_{Z^v,S,F}& \sum\limits_{v=1}^t \{\|X^v-X^vZ^v\|_F^2+\alpha \|Z^v\|_F^2+\beta w_v\|Z^v-S\|_F^2\}\\
&+\gamma Tr(F^TLF) \quad s.t. \quad Z^v\geq 0,\quad F^TF=I,
\label{obj}
\end{split}
\end{equation}
where $\alpha$, $\beta$, and $\gamma$ are regularization parameters. The objective function (\ref{obj}) enjoys the following properties: 
\begin{itemize}
\item{The last term in Eq. (\ref{obj}) functions as a regularizer on graph $S$. We tune the structure of $S$ adaptively so that we achieve the optimal condition. At the same time, it seamlessly integrates the graph construction and spectral clustering processes. }
\item{For this multi-view spectral clustering method, the graph is automatically learned from the data rather than pre-defined as in most existing spectral clustering methods. This results in a reliable and robust graph. }
\item {The graph fusion term seeks to find the underlying relationships between samples. Rather than treating each view equally, weight $w_v$ can well distinguish the different contributions of different views. Consequently, the complementary information of heterogeneous data is more effectively explored.}
\item{In this joint framework, the high-quality clustering result is utilized to guide the graph construction, which is then used to obtain a new clustering. This mutually improving approach can boost the final clustering result. }
\end{itemize}

\section{Optimization of Problem (\ref{obj})}
The variables in Eq. (\ref{obj}) are coupled to each other. We can solve them utilizing an alternating iterative strategy.

\textbf{Solving $Z^v$ when $F$ and $S$ are fixed}. The problem (\ref{obj}) becomes
\begin{equation}
\min_{Z^v} \sum\limits_{v=1}^t \{\|X^v-X^vZ^v\|_F^2+\alpha \|Z^v\|_F^2+\beta w_v\|Z^v-S\|_F^2\}.
\label{subz}
\end{equation}
We can observe that Eq. (\ref{subz}) is independent for each view. Thus, we can update $Z^v$ separately for each view. Taking the derivative of Eq. (\ref{subz}) w.r.t. $Z^v$, we have
\begin{equation*}
-2(X^v)^T(X^v-X^vZ^v)+2\alpha Z^v+2\beta w_v(Z^v-S\|_F^2).
\end{equation*}
Setting above formula to zero, we obtain
\begin{equation}
Z^v=\left((X^v)^TX^v+\alpha I+\beta w_v I\right)^{-1}\left(\beta w_v S+(X^v)^TX^v\right).
\label{solvez}
\end{equation}
\textbf{Solving $S$ when $F$ and $Z^v$ are fixed}. Remembering that $L$ is a function of $S$, thus we obtain
\begin{equation}
\min_S \sum\limits_{v=1}^t \beta w_v\|Z^v-S\|_F^2+\gamma Tr(F^TLF).
\label{objs}
\end{equation}
To solve this subproblem, we use equality 
\begin{equation*}
\sum\limits_{i,j}\frac{1}{2}\|F_{i,:}-F_{j,:}\|^2 s_{ij}=Tr(F^TLF)
\end{equation*}
and define $p_i\in\mathbb{R}^{n\times 1}$ with the $j$-th entry $p_{ij}=\|F_{i,:}-F_{j,:}\|^2$. Then problem (\ref{objs}) can be solved column-wisely 
\begin{equation}
\min_{S(:,i)} \sum\limits_{v=1}^t \beta w_v\|Z(:,i)^v-S(:,i)\|^2+\frac{\gamma}{2} p_i^TS(:,i).
\end{equation}
Its derivative w.r.t. $S(:,i)$ is $-2\beta \sum_v w_v(Z(:,i)^v-S(:,i))+\frac{\gamma}{2} p_i$, which should be zero. It yields
\begin{equation}
S(:,i)=\frac{\sum_v w_vZ^v(:,i)-\frac{\gamma p_i}{4\beta}}{\sum_v w_v}.
\label{solves}
\end{equation}
\textbf{Solving $F$ when $Z^v$ and $S$ are fixed}. It yields 
\begin{equation}
\min_F Tr(F^TLF)\quad s.t.\quad F^TF=I.
\label{solvef}
\end{equation}
The optimal solution of $F$ is obtained by the $k$ eigenvectors of $L$ corresponding to the $k$ smallest eigenvalues.

The details of solving the problem in Eq. (\ref{obj}) is summarized in Algorithm 1. We stop our algorithm if the maximum iteration number 200 is reached or the relative change of $S$ is less than $10^{-3}$. The complete implementation package is available \footnote{https://github.com/sckangz/GFSC}.
\subsection{Computational Analysis}
The main computation demand of Algorithm 1 is due to the update of $Z^v$ and $F$. Specifically, updating $Z^v$ costs about $O(n^3)$ due to the matrix inversion and multiplication. The complexity of updating $F$ is also $O(n^3)$ due to the employment of SVD operation. To make our algorithm more efficient, several off-the-shell acceleration algorithms could be utilized, e.g., skinny SVD \cite{zhang2014fast}, sampling-based methods \cite{zhang2016sampling,xu2018improved,jia2017nystrom}. In our experiments, we don't apply these acceleration techniques.

\begin{algorithm}[!tb]
%\small
\caption{The algorithm of GFSC }
\label{alg1}
{\bfseries Input:} Data matrices: $X^1,\cdots, X^t$, parameters $\alpha>0$, $\beta>0$, $\gamma>0$.\\
{\bfseries Output:} $Z^1,\cdots, Z^t$, $S$, $F$.\\
{\bfseries Initialize:} Random matrices $S$ and $F$, $w_v=1/t$.\\
{\bfseries REPEAT}
\begin{algorithmic}[1]
\STATE Update $Z^v$ according to Eq. (\ref{solvez}) for each view.
\STATE For each element $Z_{ij}^v=max(Z_{ij}^v, 0)$.
\STATE Update $S$ according to Eq. (\ref{solves}).
\STATE Update $F$ by solving the problem (\ref{solvef}).
\STATE Update $w_v$ according to (\ref{weightw}).
\end{algorithmic}
\textbf{ UNTIL} {stopping criterion is met.}
\end{algorithm}

\begin{table*}[!htbp]
\begin{center}
\renewcommand{\arraystretch}{1.}
\caption{Information of the data sets (\#Feature).\label{data}}
\label{datasets} \scalebox{.7}{
\begin{tabular}{cllll}
\hline%{1.0pt}%
%\hline
{\#View} & {BBC} &{Reuters} & {Digits} & {Caltech20} \\\hline
1& Segment1 (4659) & English (2000) &Profile correlations (216) & Gabor (48)\\
2& Segment2 (4633) & French (2000) &Fourier coefficients (76) & Wavelet moments (40)\\
3& Segment3 (4665) & German (2000) & Karhunen coefficients (64) &CENTRIST (254)\\
4& Segment4 (4684) & Spanish (2000) & Morphological (6) &HOG (1984)\\
5& -- & Italian (2000) &Pixel averages (240) & GIST (512) \\
6& -- & -- & Zernike moments (47) &LBP (928)\\\hline
\#Sample & 145 & 1200 & 2000 & 2386\\
\#Class & 2 & 6 & 10 & 20 \\
%\Xhline{1.0pt}%
\hline
\end{tabular}}
\end{center}
\end{table*}

\section{Experiments}
%In this section, we report the experimental results. 
\subsection{Data Set Descriptions}
We employ four widely used multi-view data sets for performance evaluation, namely BBC, Reuters\footnote{http://archive.ics.uci.edu/ml/datasets.html}, Digits, Caltech20\footnote{http://www.vision.caltech.edu/Image Datasets/Caltech101/}. Among them, BBC and Reuters are text data sets; Digits and Caltech20 are image data. In these cases, $Z^v$ represents the similarity between different documents or images. Table \ref{data} shows the concrete information of the data sets. According to~\cite{cai2013multi}, we normalize the data sets so that all the values of each view are in the range [-1, 1]. 
\subsection{Evaluation Metrics}
We evaluate the performance using three popular metrics: accuracy (Acc), normalized mutual information (NMI), purity \cite{peng2018integrate}. %For them, higher scores indicate better performance. One could refer to \cite{kang2017kernel} for more details of these metrics.
\begin{itemize}
\item{\textbf{Accuracy (Acc).} Accuracy is applied to find the one-to-one relationship between clusters and classes and evaluates how many data points are contained in each cluster that are from the corresponding class. It is the summation of the whole matching degree between all pair class-clusters.
\begin{equation}
Acc=\frac{1}{n}\text{max}(\sum\limits_{X_i,Y_j}N(X_i,Y_j)),
\end{equation}
where $X_i$ represents the $i$-th cluster, $Y_j$ denotes the $j$-th class, and $N(X_i,Y_j)$ denotes the number of points that are assigned to cluster $i$ but belongs to class $j$. Accuracy is defined as the maximum sum of 
$N(X_i,X_j)$ over all pairs of clusters and classes.
}
\item{\textbf{Normalized Mutual Information (NMI).} Let $X$ and $Y$ be two random variables, $H(X)$ and $H(Y)$ are their corresponding entropies. Then the NMI is defined as
\begin{equation}
NMI(X,Y)=\frac{I(X,Y)}{\sqrt{H(X)H(Y)}},
\end{equation}
where ${I(X,Y)}$ denotes the mutual information between $X$ and $Y$. Higher value indicates better performance.
}
\item{\textbf{Purity.} Purity is defined as the percent of the total number of points that are classified correctly. Then,
\begin{equation}
Purity=\frac{1}{n}\sum\limits_{i=1}^{k}\text{max}|g_i\cap t_j|,
\end{equation}
where $g_i$ denotes a cluster and $t_j$ represents the classification that has the maximum count for cluster $g_i$. 
.}
\end{itemize}
\subsection{Comparison Algorithms}
We compare with both single view and multi-view clustering algorithms.
\begin{itemize}[noitemsep]
\item{Spectral clustering (\textbf{SC}) \cite{ng2002spectral}}: We include the classic SC method as baseline method. We apply SC on each view of features. SC(1) means the implementation of SC on the 1st view. SC(Ave) means that the result is based on the average graph of views. Note that all graphs are learned from data according to Eq. (\ref{single}). 
\item{K-means clustering (\textbf{KM}): We conduct KM on the concatenated features. That is to say, we assume that all the views are of the same importance to the clustering task. }
\item{Co-training multi-view spectral clustering (\textbf{Co-train})~\cite{kumar2011cotrain}: It utilizes the eigenvector from one view to guide the graph construction in another view. Consequently, the clusterings of multiple views tend towards consensus.}
\item{Co-regularized multi-view spectral clustering (\textbf{Co-reg})~\cite{kumar2011co}: This method employs co-regularization technique to make the clusterings in different views agree with each other.}
\item{Multi-view kernel K-means (\textbf{MVKKM})~\cite{tzortzis2012kernel}: This method transforms each view into a kernel matrix and learns a weighted combination of kernels. At the same time, kernel k-means algorithm is applied to obtain the final result.}
\item{Robust multi-view K-means clustering (\textbf{RMKMC})~\cite{cai2013multi}: It adopts $\ell_{21}$-norm in traditional k-means algorithm to deal with data outliers. In addition, a weight factor is introduced for each view. }
\item{Multi-view clustering with self-paced learning (\textbf{MSPL})~\cite{xu2015multi}: This method applies the self-paced learning strategy to multi-view clustering. Hence the multi-view model is learned from easy to complex examples/views which are determined by a probabilistic smoother weighting scheme.}
\item{Auto-weighted multiple graph learning (\textbf{AMGL})~\cite{nie2016parameter}: It extends the spectral clustering method to multi-view situation. Different from our approach, the graphs are learned by adaptive neighbors approach. }
\item{Multi-view subspace clustering (\textbf{MVSC}) \cite{gao2015multi}: Multiple graphs are learned and they share the same cluster indicator matrix. Unlike our approach, there is no graph fusion process.}
\item{Diversity-induced multi-view subspace clustering (\textbf{DiMSC}) \cite{cao2015diversity}: Multiple graphs are learned and their average is inputted to the spectral clustering algorithm. Moreover, the Hilbert Schmidt Independence Criterion (HSIC) is incorporated as a diversity regularizer to explore the complementarity of multiple views.}
\item{Iterative based multi-view spectral clustering (\textbf{IMVSC}) \cite{wang2016iterative}: This method learns multiple graphs and each one is assumed to be low-rank and sparse. In addition, Laplacian regularization and views agreement are imposed on the graphs. Finally, the average of learned graphs is used for spectral clustering. }
\item{Our proposed \textbf{GFSC}. Both graph fusion and graph structure are considered in our approach. After obtaining $F$, we implement K-means on it to obtain the final discrete cluster labels. Furthermore, to see the effect of graph structure, we also compare with the approach based on problem (\ref{firstobj}) referred as \textbf{GF}. Unlike GFSC in problem (\ref{obj}), we implement the spectral clustering method separately after obtaining $S$ in GF. }
\end{itemize}

\subsection{Results}
For those methods with parameters, we tune them to achieve the best performance. For example, the range for our method is displayed in Figure \ref{sensitivity}. We repeat each algorithm 10 times and report their mean and standard deviation (std) values in Tables \ref{bbc}-\ref{caltech}. The best results are marked in boldface. According to these results, we can draw the following conclusions.
\begin{itemize}%[noitemsep]

\item{Comparing the SC performance on different views, we can see that different views indeed produce different results. This confirms the heterogeneity of multiple views. Therefore, it is essential to differentiate views when we build a multi-view learning model, just as we do in this paper.}
\item{Comparing SC(Ave) with each individual view results, we can see that naively taking the average of graphs might deteriorate the performance. In order to obtain reliable results, it is eager to design a graph fusion mechanism. }
\item{With respect to SCs and SC(Ave), our proposed GFSC method often shows better performance. This is largely due to the fact that a more accurate graph is learned in our approach. Remember that we employ both graph fusion and weighting strategy in our model.}
\item{GFSC always performs better than GF. This fully demonstrates the importance of considering the graph structure. Additionally, GF often outperforms SC(Ave). This shows the advantage of graph fusion. }
\item{Our GFSC method consistently outperforms k-means based multi-view methods, i.e., KM, MVKKM, RMKMC, MSPL. This validates the superiority of spectral clustering method. It is well-known that spectral clustering often performs better than k-means technique.}
\item{In addition, GFSC consistently performs better than AMGL. AMGL is based on adaptive neighbors which captures the local structure of data. By contrast, our graph learning is based on self-expressiveness which is supposed to grasp the global structure of data. }
\item{Our method significantly outperforms classic multi-view methods Co-train and Co-reg. Co-train and Co-reg methods construct graphs manually and they mainly regularize the multiple partitions. }
\item{Compared to state-of-the-art multi-view subspace clustering algorithms, i.e., DiMSC, MVSC, and IMVSC, our method beats them in most cases in terms of Acc, NMI, and Purity. Though they build the graphs in a similar way as ours, they don't use any graph fusion strategy. This fully demonstrates the efficacy of our graph fusion. }
\end{itemize} 
In summary, these observations validate the efficacy of our graph fusion and graph structure learning strategies.
\begin{table}[t]%[!htbp]
\begin{center}
\caption{Performance comparison on BBC (\%)}
\renewcommand{\arraystretch}{.9}
\label{bbc} \scalebox{1.}{
\begin{tabular}{c|c|c|c}
%\Xhline{1.0pt}
\hline
{Method} & {Acc}& {Purity} & {NMI}\\
\hline
{SC(1)}& {91.72(0.00)}& {99.31(0.00)}& {0.20(0.00)}\\
{SC(2)}& {93.79(0.00)}& {98.62(0.00)}& {13.71(0.00)}\\
{SC(3)}& {91.17(1.74)}& {98.62(2.18)}& {0.18(0.05)}\\
{SC(4)}& {91.72(0.00)}& {99.31(0.00)}& {0.20(0.00)}\\
{SC(Ave)}&91.72(0.00)& 99.31(0.00)& 0.20(0.00)\\
{KM}& 91.59(0.31)& 90.24(0.24) & 14.10(1.30)\\

{Co-train}& {91.27(0.00)}& {87.57(1.20)} & {3.50(0.00)}\\

{Co-reg}& {90.90(0.76)}& {90.78(1.40)} & {6.8(0.30)}\\

{MVKKM}& {84.00(6.13)}& {89.01(2.35)} & {8.3(0.64)}\\

{RMKMC}& {91.31(0.62)}& { 89.67(1.80)} & {8.00(0.74)}\\

{MSPL} & 80.41(13.24) & 90.41(0.00) & 10.11(9.48) \\

{AMGL}& {89.66(0.00)}& {91.00(0.67)} & {11.2(0.00)}\\
{DiMSC}&93.79(0.00)& {94.62}(0.00) & 13.71(0.00)\\
{MVSC}& {91.03(0.00)}& {95.62}(0.00) & {0.41(0.00)}\\
%{IMVSC}& {87.59(0.00)}& {91.03}(0.67) & {37.90(0.00)}\\
{IMVSC}& {87.59(0.00)}& {91.03}(0.67) & {7.90(0.00)}\\
{GF}& {91.72(0.00)}& {99.31}(0.00) & {0.20(0.00)}\\
{GFSC}& \textbf{93.85}(8.22)& \textbf{99.42}(7.29)& \textbf{15.13}(8.45) \\
\hline
%\Xhline{1.0pt}
\end{tabular}}
\end{center}
\end{table}

\begin{table}[t]%[!htbp]
\begin{center}
\renewcommand{\arraystretch}{.9}
\caption{Clustering performance on Reuters (\%)}
\label{Reuters} \scalebox{1.0}{
\begin{tabular}{c|c|c|c}
%\Xhline{1.0pt}
\hline
{Method} & {Acc}& {Purity} & {NMI}\\
\hline
{SC(1)}& {42.98(3.82)}& {60.09(4.49)} & {23.48(2.74)}\\
{SC(2)}& {42.67(2.22)}& {65.79(6.09)} & {25.06(2.07)}\\
{SC(3)}& {40.76(3.84)}& {59.29(5.63)} & {21.53(2.60)}\\
{SC(4)}& {43.43(2.43)}& {65.33(6.72)} & {25.04(1.05)}\\
{SC(5)}& {40.98(3.45)}& {60.39(6.02)} & {21.95(2.51)}\\
{SC(Ave)}&44.44(4.01)& 60.35(5.52)& 25.19(2.48)\\
{KM}& 24.57(4.52) & 25.48(4.37) & 11.78(5.01)\\

{Co-train}& 17.00(0.10) & 17.15(0.07) & 9.40(0.11)\\

{Co-reg}& 20.62(1.24) & 20.95(1.32) & 2.33(0.34)\\

{MVKKM}& 20.48(3.82) & 20.65(3.83) & 5.77(3.66)\\

{RMKMC}& 22.42(6.54) & 22.55(6.57) & 7.21(7.29)\\

{MSPL} & 24.87(5.98) & 28.12(4.97) & 11.50(4.28) \\

{AMGL}& 18.35(0.15) & 20.08(0.54) & 6.38(1.00)\\
{DiMSC}&39.60(1.32)& 46.28(1.74) &18.17(0.64)\\
{MVSC}& {25.08(0.39)}& {\textbf{80.11}(5.50)} & {6.60(0.68)}\\
{IMVSC}& {30.23(0.40)}& {35.73(1.16)} & {9.26(0.22)}\\
{GF}& {44.28(2.60)}& {58.36(3.23)} & {25.42(1.63)}\\
{GFSC}& \textbf{44.92}(2.68)& 59.40(2.50)& \textbf{25.73}(2.52)\\
\hline
%\Xhline{1.0pt}
\end{tabular}}
\end{center}
\end{table}

\begin{table}[t]%[!htbp]
\begin{center}
\renewcommand{\arraystretch}{.9}
\caption{Performance comparison on Digits (\%)}
\label{Digit} \scalebox{1.0}{
\begin{tabular}{c|c|c|c}
%\Xhline{1.0pt}
\hline
{Method} & {Acc}& {Purity} & {NMI}\\
\hline
{SC(1)}& {62.54(4.56)}& {70.94(3.77)} & {62.65(2.39)}\\
{SC(2)}& {59.30(4.08)}& {64.21(1.24)} & {57.35(1.23)}\\
{SC(3)}& {53.01(5.57)}& {75.5(2.12)} & {55.55(3.78)}\\
{SC(4)}& {23.17(4.22)}& {89.61(2.58)} & {23.83(5.18)}\\
{SC(5)}& {30.61(4.43)}& {81.13(2.85)} & {29.39(5.32)}\\
{SC(6)}& {55.94(2.65)}& {57.77(1.53)} & {48.16(0.99)}\\
{SC(Ave)}&77.40(6.63)& 86.22(2.45)& 79.28(2.85)\\
{KM}& 54.46(5.60) & 58.64(2.92) & 58.25(0.85)\\

{Co-train}& 71.42(4.21) & 74.86(2.62) & 71.06(1.07)\\

{Co-reg}& 83.38(7.35) & 85.17(4.98) & 77.97(2.92)\\

{MVKKM}& 58.81(3.50) & 62.40(3.40) & 62.91(2.60)\\

{RMKMC}& 63.04(3.36) & 65.74(2.16) & 66.57(1.18)\\

{MSPL} & 68.00(1.12) & 68.99(1.17) &70.42(1.95) \\

{AMGL}& 73.61(10.29) & 76.48(8.54) & 81.86(4.53)\\
{DiMSC}&42.72(1.94)& 45.65(0.97) & 37.89(0.87)\\
{MVSC}& {79.60(2.54)}& {87.19(1.48)} & {73.89(1.93)}\\
{IMVSC}& {71.03(0.65)}& {73.95(4.24)} & {67.20(2.88)}\\
{GF}& {87.76(5.32)}& {89.44(2.21)} & {83.28(2.47)}\\
{GFSC}& \textbf{89.45}(5.10)& \textbf{91.38}(1.03)& \textbf{85.37}(1.96)\\
\hline
%\Xhline{1.0pt}
\end{tabular}}
\end{center}
\end{table}

\begin{table}[t]%[!htbp]
\begin{center}
\renewcommand{\arraystretch}{.9}
\caption{Performance comparison on Caltech20 (\%)}
\label{caltech} \scalebox{1.0}{
\begin{tabular}{c|c|c|c}
%\Xhline{1.0pt}
\hline
{Method} & {Acc}& {Purity} & {NMI}\\
\hline
{SC(1)}& {33.82(0.00)}& {\textbf{99.20}(0.00)}& {12.89(0.00)}\\
{SC(2)}& {34.18(2.54)}& {97.91(3.76)}& {2.34(3.40)}\\
{SC(3)}& {49.80(5.61)}& {85.28(3.48)}& {19.71(4.50)}\\
{SC(4)}& {53.13(4.77)}& {66.05(4.81)}& {61.03(2.13)}\\
{SC(5)}& {33.65(0.03)}& {\textbf{99.20}(0.01)}& {1.14(0.00)}\\
{SC(6)}& {57.36(1.02)}& {80.72(4.33)}& {31.22(1.37)}\\
{SC(Ave)}&65.19(1.17)& 86.97(0.58)& 45.28(6.19)\\
{KM}& 31.40(1.30) & 60.06(0.38) & 37.05(0.41)\\

{Co-train}& 38.94(2.10) & 69.77(1.42) & 50.90(1.12)\\

{Co-reg}& 34.38(0.79) & 65.59(1.03) & 46.42(0.96)\\

{MVKKM}& 44.87(2.49) & 72.84(0.72) & 54.06(1.23)\\

{RMKMC}& 33.35(1.47) & 64.22(0.89) & 42.44(0.67)\\

{MSPL} & 33.49(0.00) & 34.24(0.00) & 35.80(0.00) \\

{AMGL}& 52.28(2.91) & 67.60(2.31) & 56.61(1.93)\\
{DiMSC}&33.89(1.45)& 37.78(1.35) & 39.33(1.16)\\
{MVSC}& {44.96(2.06)}& {50.87(2.35)} & {45.36(0.88)}\\
{IMVSC}& {42.07(1.95)} & {46.19(1.81)}& {51.18(0.90)}\\
{GF}& {66.95(1.90)} & {79.50(4.28)}& {56.19(3.07)}\\
{GFSC}& \textbf{70.24}(2.94)& 81.49(1.88)& \textbf{63.09}(2.49)\\
\hline
%\Xhline{1.0pt}
\end{tabular}}
\end{center}
\end{table}
\begin{figure*}[!htb]
\centering%[$\alpha=10$\label{ba}]
\subfloat[$\gamma=10^{-7}$]{\includegraphics[width=.48\textwidth]{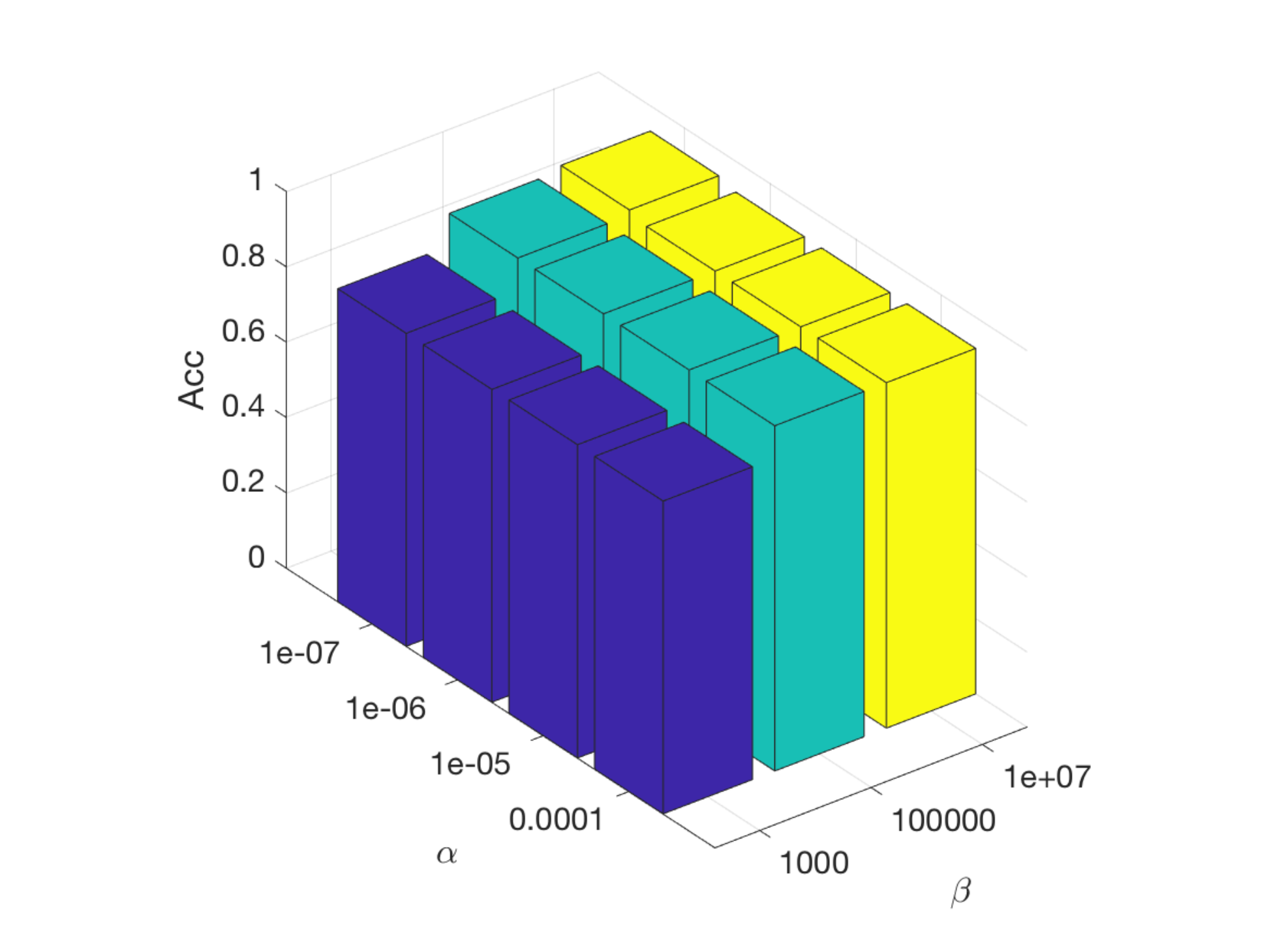}}
\subfloat[$\gamma=10^{-5}$]{\includegraphics[width=.48\textwidth]{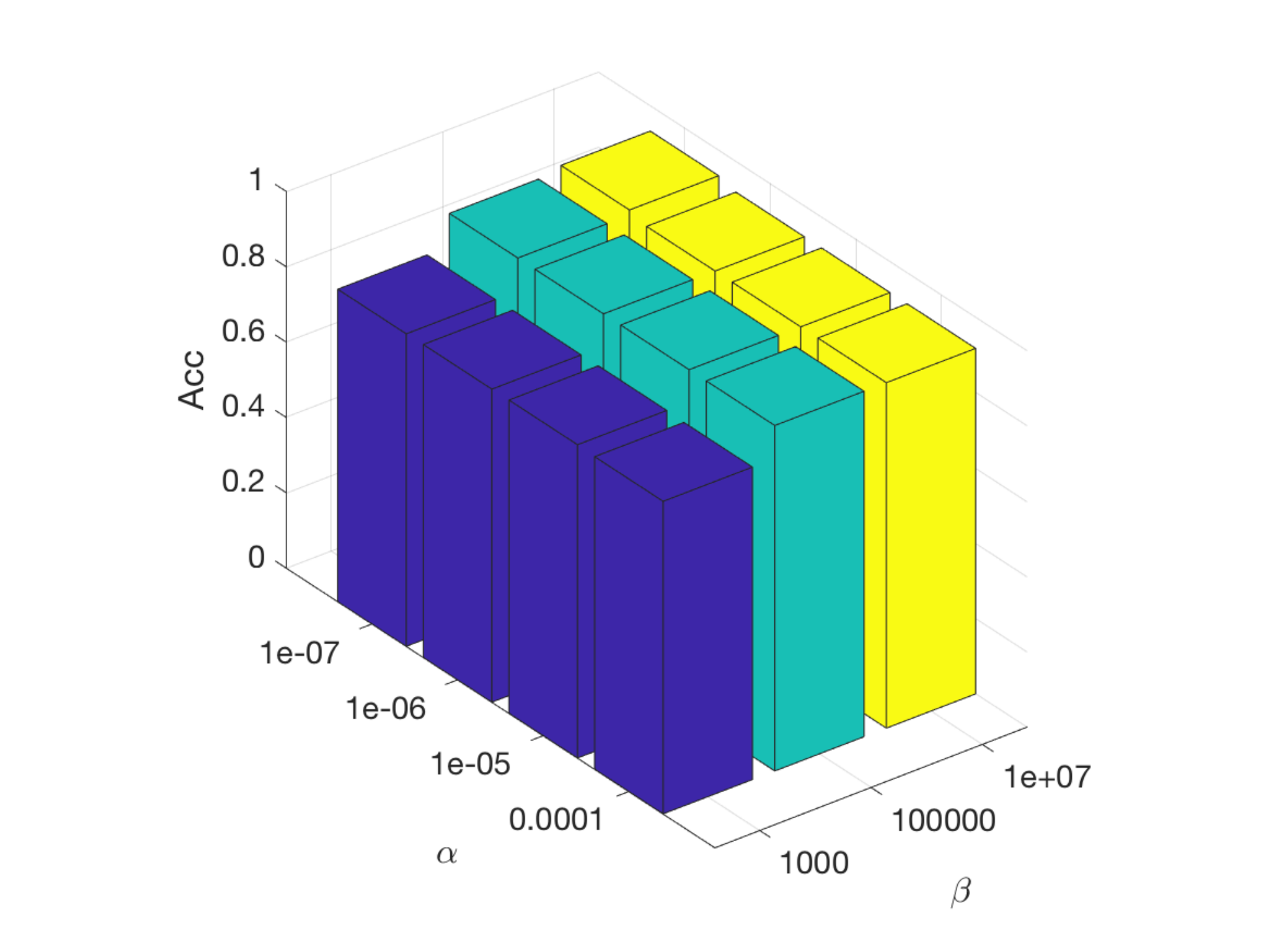}}
\caption{Acc w.r.t. $\alpha$, $\beta$, and $\gamma$ on BBC data.}
\label{bbc}
\end{figure*}

\begin{figure*}[!htb]
\centering%[$\alpha=10$\label{ba}]
\subfloat[$\gamma=0.01$]{\includegraphics[width=.48\textwidth]{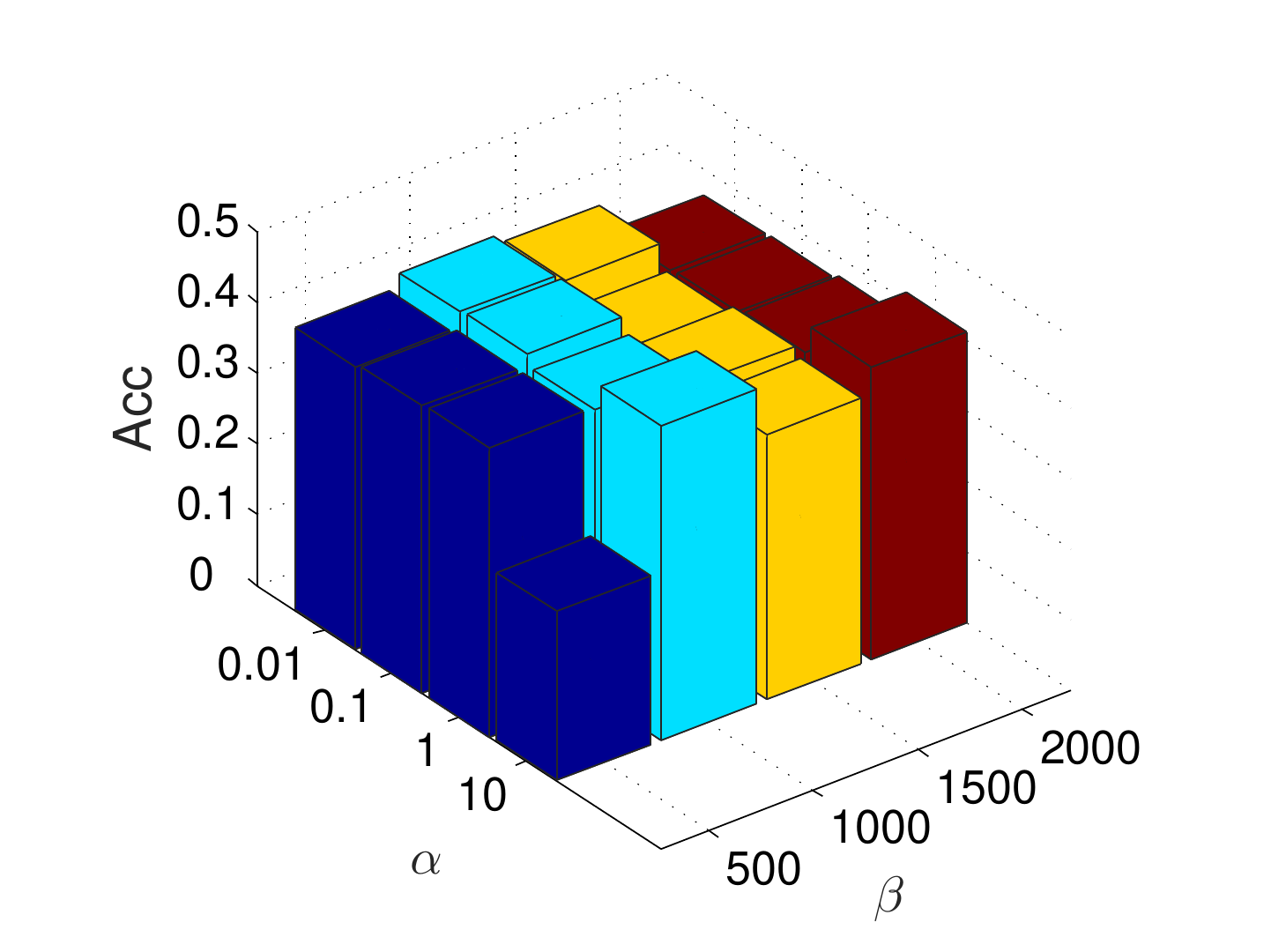}}
\subfloat[$\gamma=0.1$]{\includegraphics[width=.48\textwidth]{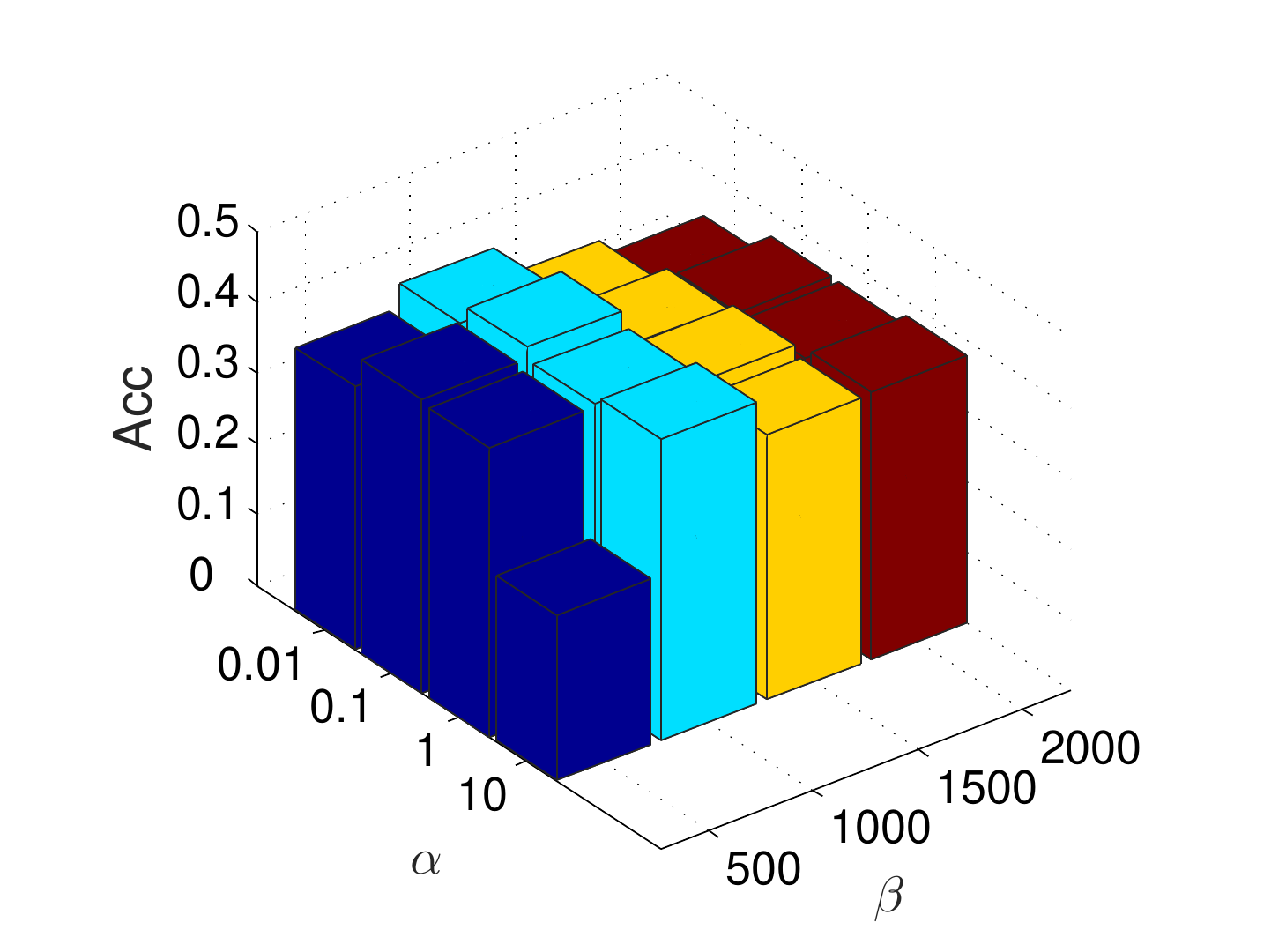}}
\caption{Acc w.r.t. $\alpha$, $\beta$, and $\gamma$ on Reuters data.}
\label{reuters}
\end{figure*}

\begin{figure*}[!htb]
\centering%[$\alpha=10$\label{ba}]
\subfloat[$\gamma=0.0001$]{\includegraphics[width=.48\textwidth]{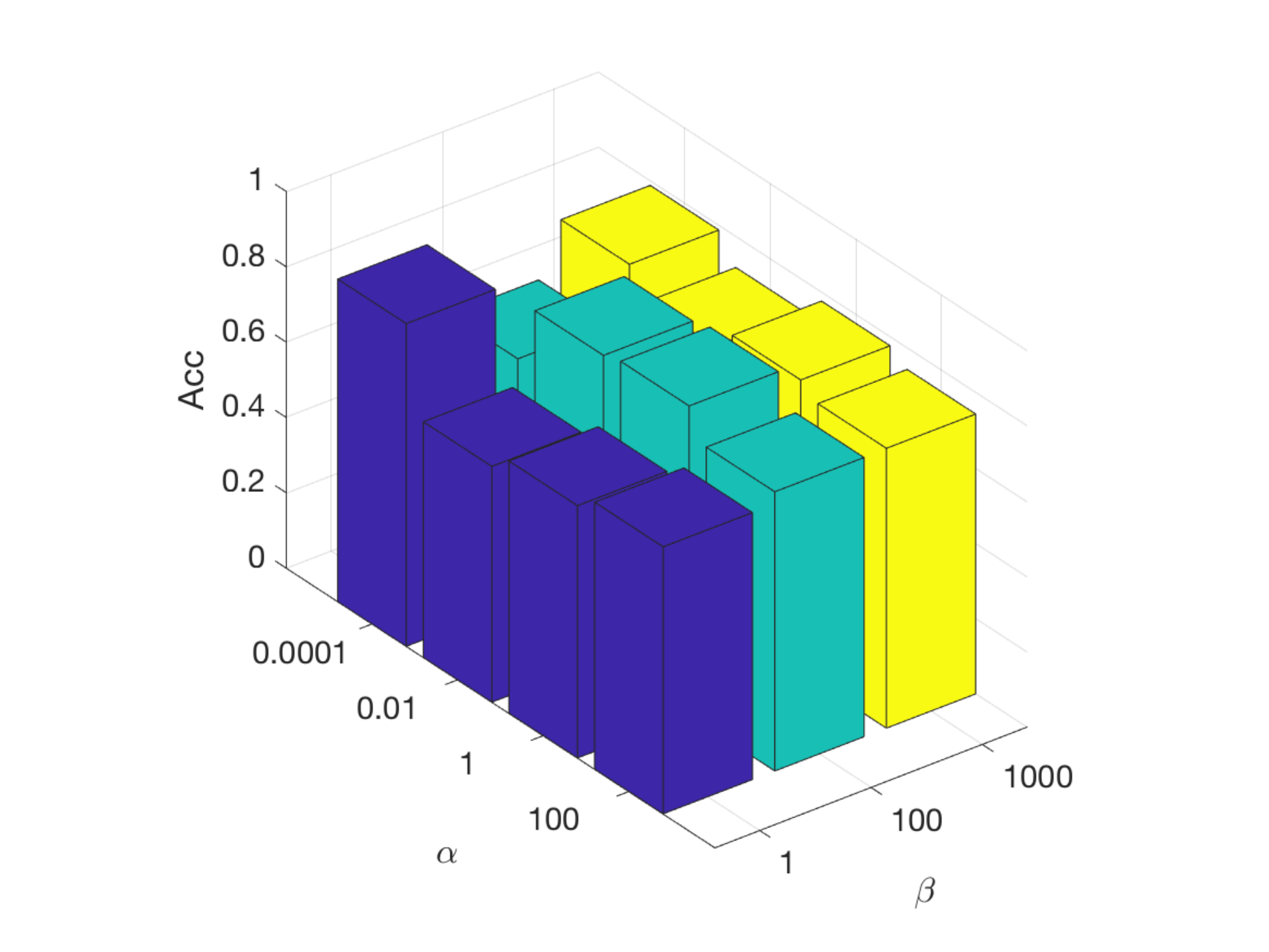}}
\subfloat[$\gamma=0.001$]{\includegraphics[width=.48\textwidth]{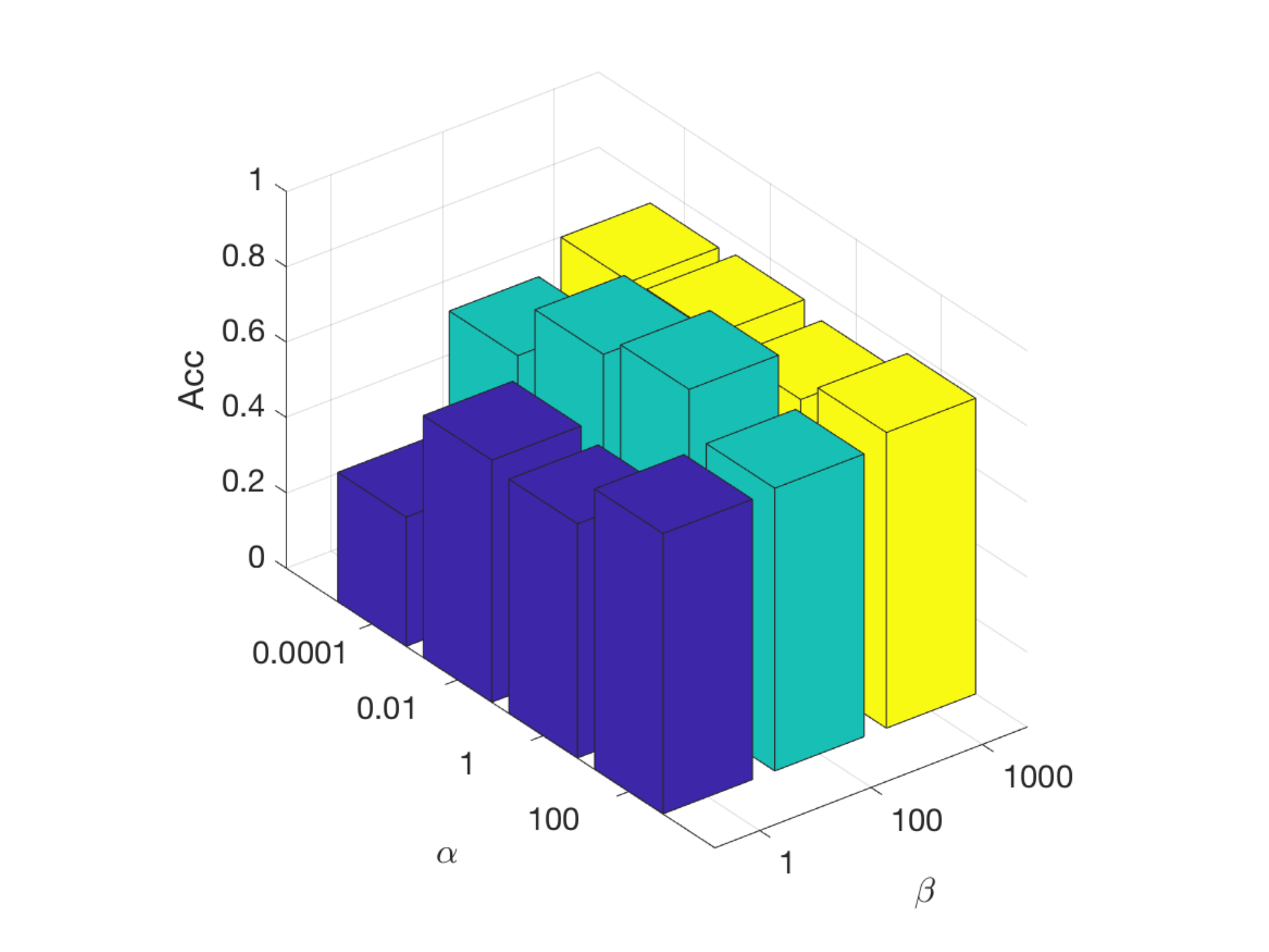}}
\caption{Acc w.r.t. $\alpha$, $\beta$, and $\gamma$ on Digits data.}
\label{digits}
\end{figure*}

\begin{figure*}[!htb]
\centering%[$\alpha=10$\label{ba}]
\subfloat[$\gamma=10^{-7}$]{\includegraphics[width=.48\textwidth]{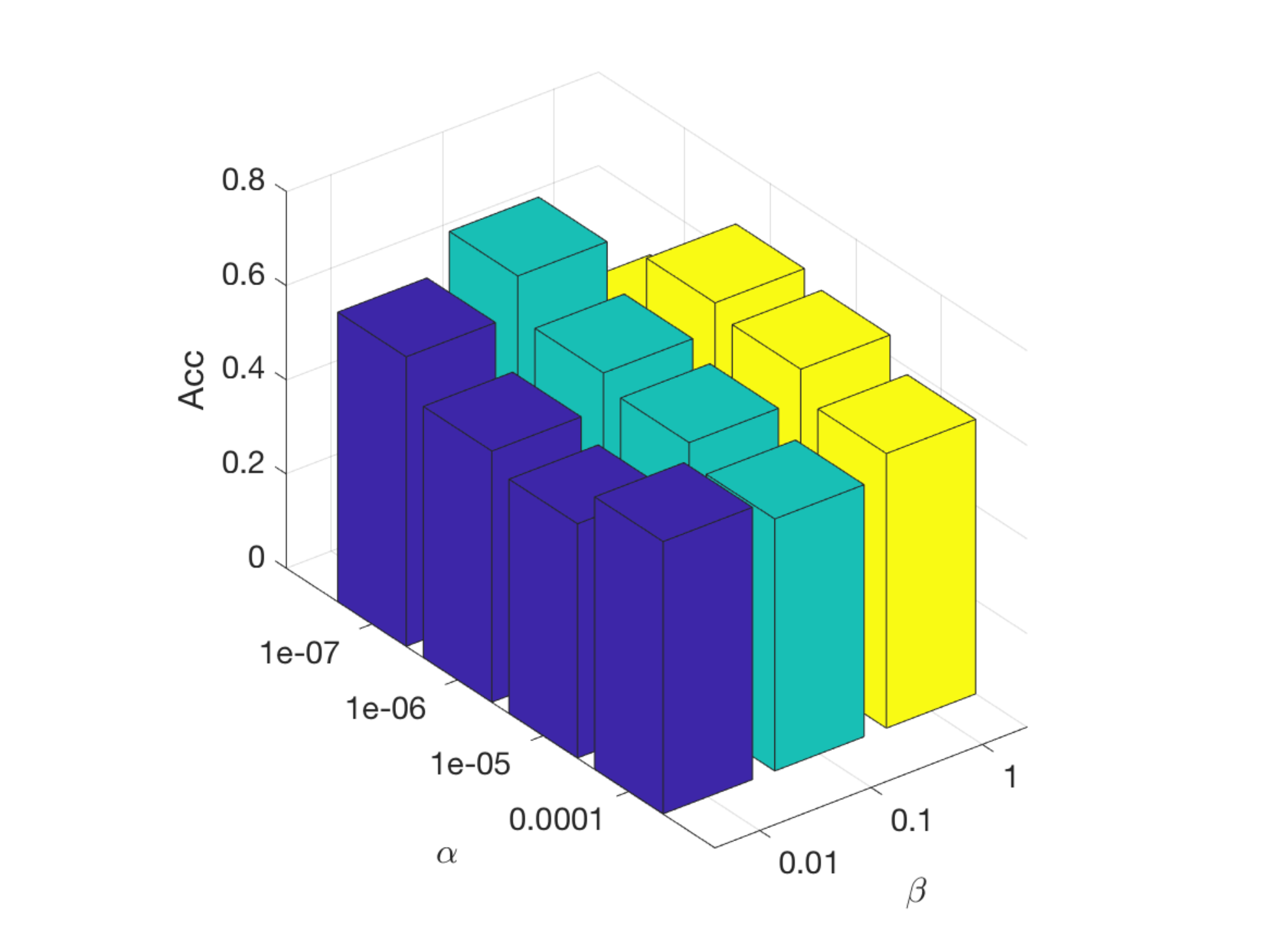}}
\subfloat[$\gamma=10^{-5}$]{\includegraphics[width=.48\textwidth]{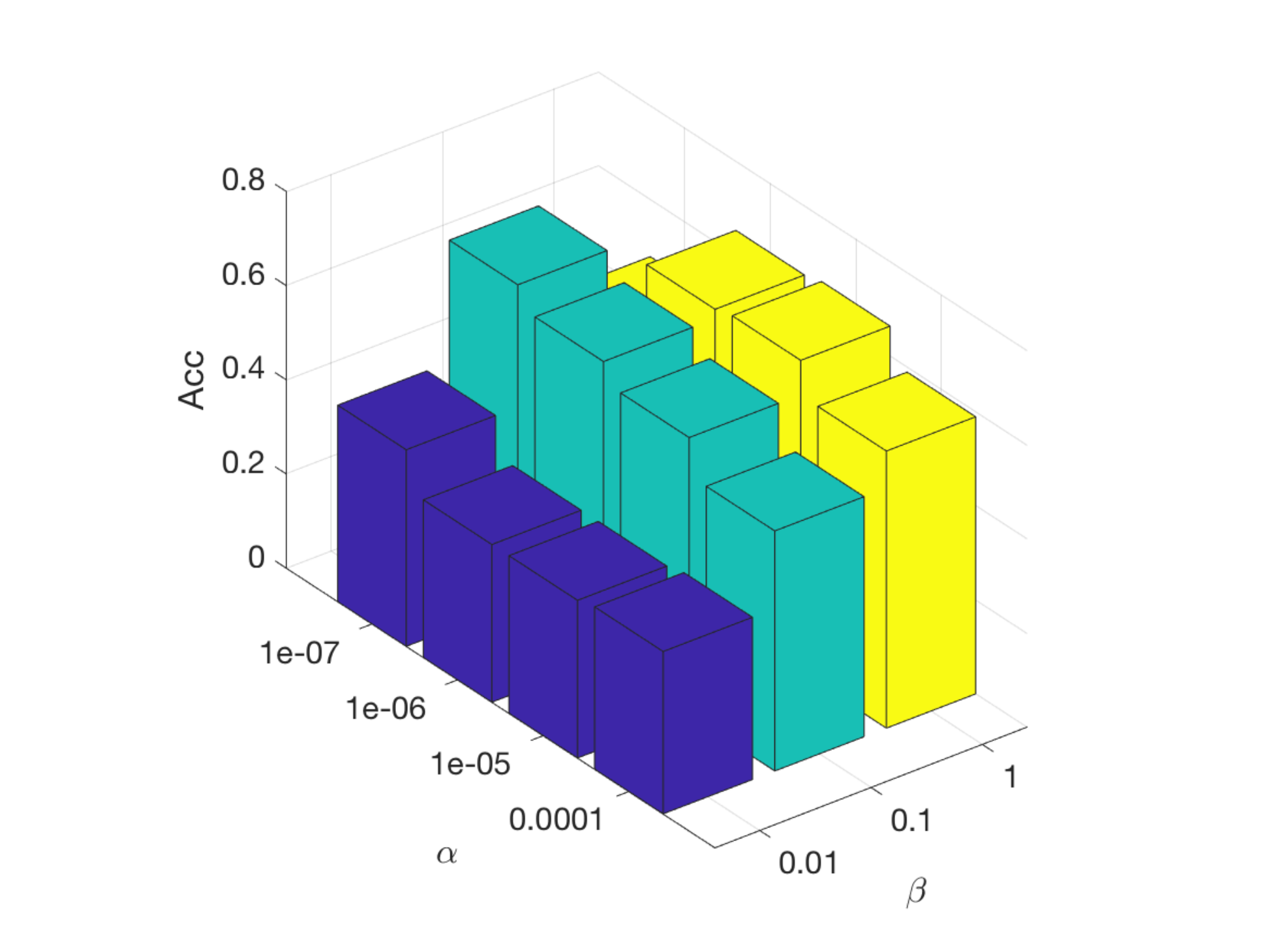}}
\caption{Acc w.r.t. $\alpha$, $\beta$, and $\gamma$ on Caltech20 data.}
\label{reuters}
\end{figure*}
\subsection{Parameter Analysis}
In our proposed model, there are three parameters $\alpha$, $\beta$, and $\gamma$ that need to be set properly. We choose their values by grid searching. Figures \ref{bbc}-\ref{caltech} show the range for each dataset and the sensitivity of the accuracy with regard to the parameters. As can be seen, the optimal parameters are $\{10^{-7},10^{7},10^{-5}\}$, $\{10, 10^{3}, 0.01\}$,$\{10^{-4},1,10^{-4}\}$,$\{10^{-7},0.1,10^{-7}\}$ for BBC, Reuters, Digits, Caltech20, respectively. Overall, our method performs stably to some extent w.r.t. a wide range of parameter values. %This conclusion is also true for other measures.

\section{Conclusion}
In this paper, we proposed a novel multi-view spectral clustering method. Unlike many existing methods, which often use averaged graph to perform spectral clustering, we propose a way to fuse graphs to achieve a consensus graph. A parameter-free weighting scheme is introduced to distinguish the contributions of different graphs. Moreover, the cluster structure of the consensus graph is also considered in the proposed method. Consequently, the proposed approach integrates graph learning, fusion, and spectral clustering into a unified framework. These three subtasks are mutually boosted based on an alternating iterative optimization strategy. Experiments on benchmark data sets verify the effectiveness of the proposed methods. The results show that both the 
consensus graph and the graph structure help improve the clustering quality.

\section{Acknowledgement}
This paper was in part supported by Grants from the Natural
Science Foundation of China (Nos. 61806045, 61572111, and 61772115), two Fundamental
Research Fund for the Central Universities of China (Nos. ZYGX2017KYQD177 and
A03017023701012), and a 985 Project
of UESTC (No. A1098531023601041).
\section{References}
%\vfill\eject
\bibliographystyle{elsarticle-num}
\bibliography{ref}

\end{document}